\documentclass[10pt,twocolumn,letterpaper]{article}

\usepackage[pagenumbers]{cvpr} 

\usepackage{caption}
\usepackage{graphicx}
\usepackage{amsmath}
\usepackage{amssymb}
\usepackage{booktabs}
\usepackage{mathalias}
\usepackage{array}
\usepackage[normalem]{ulem}

\usepackage[pagebackref,breaklinks,colorlinks]{hyperref}

\usepackage[capitalize]{cleveref}
\crefname{section}{Sec.}{Secs.}
\Crefname{section}{Section}{Sections}
\Crefname{table}{Table}{Tables}
\crefname{table}{Tab.}{Tabs.}

\usepackage{comment} 

\usepackage{color}
\definecolor{darkorange}{rgb}{1.0, 0.55, 0.0}

\def\cvprPaperID{2345} 
\def\confName{CVPR}
\def\confYear{2022}

\begin{document}

\title{Playable Environments: Video Manipulation in Space and Time}

\author{Willi Menapace\thanks{This work was partially done while interning at MPI for Informatics}\\
\normalsize University of Trento \\
\and
St\'{e}phane Lathuili\`{e}re\thanks{Equal senior contribution}\\
\normalsize LTCI, T\'{e}l\'{e}com Paris\\
\normalsize Institut Polytechnique de Paris
\and
Aliaksandr Siarohin\\
\normalsize University of Trento
\and
Christian Theobalt\footnotemark[2]\\
\normalsize MPI for Informatics, SIC
\and
Sergey Tulyakov\footnotemark[2]\\
\normalsize Snap Inc.
\and
Vladislav Golyanik\footnotemark[2]\\
\normalsize MPI for Informatics, SIC
\and
Elisa Ricci\footnotemark[2]\\
\normalsize University of Trento \\
\normalsize Fondazione Bruno Kessler
}


\maketitle

\begin{abstract}

We present Playable Environments---a new representation for interactive video generation and manipulation in space and time. With a single image at inference time, our novel framework allows the user to move objects in 3D while generating a video by providing a sequence of desired actions. The actions are learnt in an unsupervised manner. The camera can be controlled to get the desired viewpoint. Our method builds an environment state for each frame, which can be manipulated by our proposed action module and decoded back to the image space with volumetric rendering. 
To support diverse appearances of objects, we extend neural radiance fields with style-based modulation. 
Our method trains on a collection of various monocular videos requiring only the estimated camera parameters and 2D object locations. To set a challenging benchmark, we introduce two large scale video datasets with significant camera movements. As evidenced by our experiments, playable environments enable several creative applications not attainable by prior video synthesis works, including playable 3D video generation, stylization and manipulation\footnote{\href{https://willi-menapace.github.io/playable-environments-website/}{willi-menapace.github.io/playable-environments-website}}.
\end{abstract}
\vspace{-0.5cm}

\section{Introduction}

\setlength{\belowcaptionskip}{0pt}
\begin{figure}
    \centering
    \includegraphics[trim=0 0 0 14, clip,width=\columnwidth]{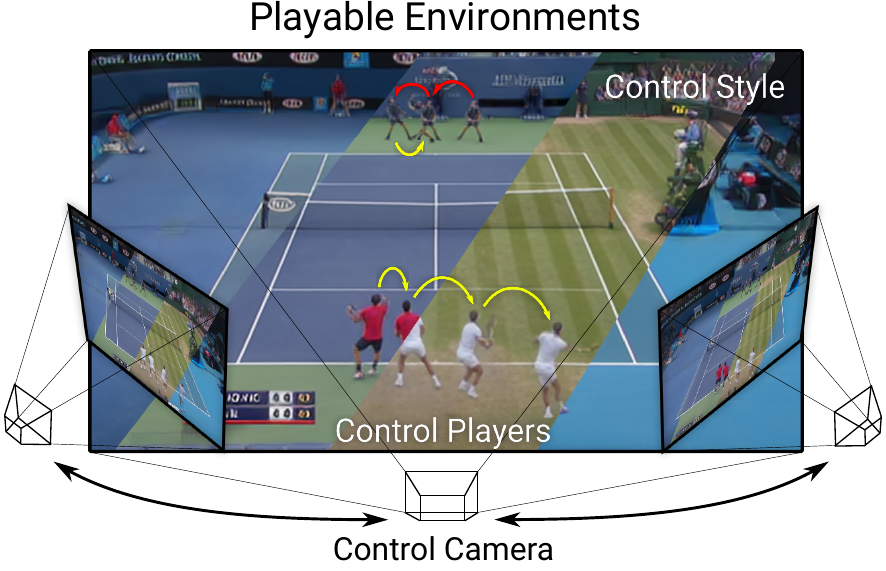}

    \caption{
    Given a single initial frame, our method creates \emph{playable environments} that allow the user to interactively generate different videos by specifying discrete actions to control players, manipulating the camera trajectory and indicating the style for each object in the scene.}
    \label{fig:teaser}
    \vspace{-0.3cm}
\end{figure}

What would you change in the last tennis match you saw? The actions of the player? The style of the field, or, perhaps, the camera trajectory to observe a highlight more dramatically? To do so interactively, 
the geometry and the style of the field and the players need to be reconstructed. Players' actions need to be understood and the outcomes of future actions anticipated. To enable these features one needs to reconstruct the observed \emph{environment} in 3D and provide simple and intuitive interaction, offering an experience similar to \emph{playing} a video game. We call these representations Playable Environments (PE). 

Such a representation enables multiple creative applications, such as 3D- and action-aware video editing, camera trajectory manipulation, changing the action sequence, the agents and their styles, or continuing the video in time, beyond the observed footage. Fig. \ref{fig:teaser} shows a playable environment for tennis matches. In it, the user specifies actions to move the players, controls the viewpoint and changes the style of the players and the field. The environment can be played, akin to a video game, but with real objects.
%

In this work, we propose a method to construct PEs of complex scenes that supports a large set of interactive manipulations.
Trained on a dataset of monocular videos, our method presents six core characteristics listed in Tab.~\ref{table:requirements} that enable the creation of such PEs.
Our framework allows the user to interactively generate videos by providing discrete actions $\langle$\textbf{1}$\rangle$ and controlling the camera pose $\langle$\textbf{2}$\rangle$. Furthermore, it can represent environments with multiple objects $\langle$\textbf{3}$\rangle$ with varying poses $\langle$\textbf{4}$\rangle$ and appearances $\langle$\textbf{5}$\rangle$ and is robust to imprecise inputs $\langle$\textbf{6}$\rangle$. In particular, we do not require ground-truth camera intrinsics and extrinsincs, but assume they can be estimated for each frame. Neither do we assume ground-truth object locations, but rely on an off-the-shelf object detector \cite{ren2015faster} to locate the agents in 2D, such as both tennis players. No other supervision is required.


Playable Environments encapsulate and extend representations built by several prior image or video manipulation methods.
Novel view synthesis and volumetric rendering methods support re-rendering of static scenes. However, while some methods support moving or articulated objects \cite{pumarola2021dnerf,tretschk2021nonrigid,Ost_2021_CVPR,yuan2021star}, it is challenging for them to handle dynamic environments and they do not allow user interaction, making them undesirable for modeling compelling environments.
Video synthesis methods manipulate videos by predicting future frames \cite{lee2018savp,kumar2020videoflow,tulyakov2018moco,tian2021good}, animating~\cite{siarohin2019monkeynet,Siarohin2019firstorder,siarohin2021motion} or playing videos \cite{menapace2021pvg}, but environments modeled with such methods typically lack camera control and multi-object support. Consequently, these methods limit interactivity as they do not take into account the 3D nature of the environment. 

\begin{table}
\begin{center}
\setlength{\tabcolsep}{1.8pt}
\footnotesize

\begin{tabular}{lp{0.08\textwidth}p{0.35\textwidth}}
\toprule
\bf  &\bf Name& \bf Description\\
\midrule
$\langle$\textbf{1}$\rangle$ &\textit{Playability} & The user can control generation with discrete actions.\\
$\langle$\textbf{2}$\rangle$ & \textit{Camera control} & The camera pose is explicitly controlled at test time.\\
$\langle$\textbf{3}$\rangle$ & \textit{Multi-object}& Each object is explicitly modeled. \\
$\langle$\textbf{4}$\rangle$& \textit{Deformable objects} & The model handles deformable object such as human bodies\\
$\langle$\textbf{5}$\rangle$ & \textit{Appearance changes} & The model handles objects whose appearance is not constant is the training set \\
$\langle$\textbf{6}$\rangle$ & \textit{Robustness} & The model is robust to calibration and localization errors.\\

\bottomrule
\end{tabular}
\end{center}
\caption{Characteristics of our method for Playable Environments. Each row is referred in the text with $\langle \cdot \rangle$ symbols.}

\label{table:requirements}
\end{table}


Our method consists of two components. The first one is the synthesis module. It extracts the state of the environment---location, style and non-rigid pose of each object---and renders the state back to the image space. Recently introduced Neural Radiance Fields (NeRFs)~\cite{mildenhall2020nerf} represent an attractive tool for their ability to render novel views. 
In this work, we introduce a style-based modification of NeRF to support objects of different appearances. 
Furthermore, we propose a compositional non-rigid volumetric rendering approach handling the rigid parts of the scene and non-rigid objects. 
The second component---the action module---enables playability. It takes two consecutive states of the environment and predicts an action with respect to the camera orientation. We train our framework using reconstruction losses in the image space and the state space, and a novel loss for action consistency. Finally, to improve temporal dynamics, we introduce a temporal discriminator that operates on sequences of environment states. 
%
%

To thoroughly evaluate $\langle$\textbf{1}$-$\textbf{6}$\rangle$, we introduce two complementary large-scale datasets for the training of playable environments, a synthetic and a real one. The first is intended to evaluate $\langle$\textbf{1}$-$\textbf{5}$\rangle$, with a particular focus on camera control thanks to the synthetic ground truth, the second to evaluate $\langle$\textbf{1}$-$\textbf{6}$\rangle$, with a particular focus on $\langle$\textbf{4}$-$\textbf{6}$\rangle$ given the high diversity present in this dataset. We propose an extensive evaluation of our method with several baselines derived from existing NeRF and video generation methods. These experiments show that our method is able to generate high-quality videos and outperforms all baselines in terms of playability, camera control and video quality.

In summary, the primary contributions of this work are as follows: \textbf{a new framework} for the creation of compelling Playable Environments with the characteristics in Tab.~\ref{table:requirements}, featuring \textbf{a new compositional NeRF} that handles deformable objects with different visual styles and an \textbf{action module} that operates in the latent space of our NeRF model; \textbf{two challenging large-scale datasets} for training and evaluating PEs to stimulate future research in this area.

\section{Related Works} 


\noindent\textbf{Video generation} has seen incredible progress over past years. The video synthesis task has numerous formulations which mostly differ in the type of conditional information that is used for generation. The generation process could be conditioned on previous frames~\cite{finn2016cdna, mathieu2015deep, vondrick2015anticipating, lee2018savp, tulyakov2018moco}, on another video~\cite{wang2018video,siarohin2019monkeynet,Siarohin2019firstorder,siarohin2021motion}, on the pose of the agent~\cite{chan2019everybody} or even be completely unconditional~\cite{saitotrain,tulyakov2018moco}. Moreover, several works proposed to condition the generation of each single frame on an action label~\cite{chiappa2017recurrent,nunes2020action, oh2015action,Kim2020_GameGan}. Still, all these methods require action supervision for training. 

\noindent\textbf{Playable video generation (PVG)} was recently introduced in Menapace \emph{et al.}~\cite{menapace2021pvg}. Differently from prior works in this domain which required annotated action labels~\cite{Kim2020_GameGan,kim2021drivegan}, their method, CADDY, automatically infers actions during training in a completely unsupervised manner from raw videos. This method is closely related to ours. However, CADDY assumes only a single controllable object while here we also model the camera movement, complex 3D interactions and support a variety of object appearances.

\noindent \textbf{Novel view synthesis} methods traditionally utilized depth maps~\cite{CDSD13,Soft3DReconstruction} or multi-view geometry~\cite{Kopf2013,Zitnick2004HighqualityVV,Seitz2006comparison} in order to reconstruct underlying 3D representation and later render new views of the corresponding scene. Recently, Neural Radiance Fields (NeRF)~\cite{mildenhall2020nerf} revolutionized the field of novel view synthesis. The main idea of NeRF~\cite{mildenhall2020nerf} is to model the scene as a continuous 5D function, usually represented by MLP, and directly query this function along the camera rays. Since the pioneering work in \cite{mildenhall2020nerf}, numerous NeRF-based models~\cite{mildenhall2020nerf} have been proposed. For instance, some works proposed to decompose the foreground and background~\cite{kaizhang2020nerfplusplus,Niemeyer2021CAMPARI}. Other works generalised NeRF~\cite{mildenhall2020nerf} to dynamic scenes~\cite{Ost_2021_CVPR,tretschk2021nonrigid,yuan2021star,zhang2021stnerf}. GIRAFFE~\cite{niemeyer2021giraffe} and GANcraft~\cite{hao2021GANcraft} proposed to utilize an internal representation that is rendered in a feature space and later decoded by a standard 2D convolutional network.
However, none of these methods is able to generalize to multiple monocular videos, several moving and deforming objects, and diverse objects and scene appearances. Comparatively, our method can be trained with such data. Moreover, for enriching the interactivity of the playable environment, our method can control objects in the scene with action labels that are discovered in an unsupervised manner.


\setlength{\abovedisplayskip}{3pt}
\setlength{\belowdisplayskip}{3pt}

\section{Method}



\begin{figure}[t]
    \centering
    \includegraphics[width=0.90\columnwidth]{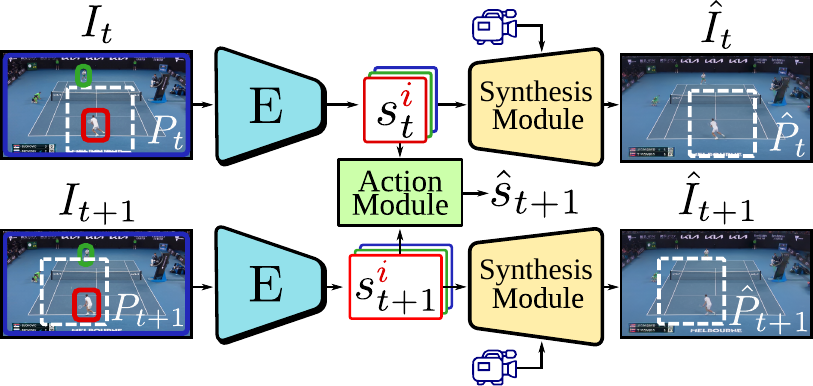}
    \caption{\textbf{Overview of our framework}. 
      The encoder $E$ extracts {environment states} for every object in the scene. The synthesis module follows a NeRF-like architecture to reconstruct the input frame and allows for camera manipulation. We introduce the action module that learns to encode state dynamics with discrete action labels. At test time, these learned action labels are provided by the user to control the generated content.} 
    \label{fig:pipeline}
\end{figure}

Our framework is based on the encoder-decoder architecture shown in Fig.~\ref{fig:pipeline} whose design is driven by the playable environment characteristics $\langle$\textbf{1}-\textbf{6}$\rangle$ in Tab.~\ref{table:requirements}.
At time $t$, the encoder network outputs {state} vector $s_t^i$ for every object $i$ in the scene. To enable playability $\langle$\textbf{1}$\rangle$, we include an action module in the bottleneck layer that has two goals. First, it learns discrete action labels in an unsupervised manner. More precisely, we learn to discretize the transition from $s_t^i$ to $s_{t+1}^i$ using an action labels $a_t^i\in\{1,...,K\}$, where the number of actions $K$ is a hyper-parameter specified before training. 
Second, 
the action module is used at test time to condition the next frame generation on the action selected by the user. Finally, the decoder network, referred to as the synthesis module, is in charge of reconstructing the input frame combining the {state} of every object and the camera parameters to allow for camera control $\langle$\textbf{2}$\rangle$.
The synthesis and action modules are trained in two separate phases using reconstruction as the main driving loss.

To handle environments with multiple objects $\langle$\textbf{3}$\rangle$, we adopt a compositional formulation for our encoder-decoder: we decompose the environment into a predefined set of objects. We distinguish between two object categories, namely static objects (\eg background) and playable objects (\eg human), where the latter are the dynamic objects the user will be able to control. We define the {environment state} of object $i$ as $s_t^i=(x_t^i, w_t^i, \pi_t^i)$ where $x_t^i$ is the position of the object in the environment, $w^i$ is a style descriptor, and $\pi^i$ is the object pose. We introduce $w^i$ and $\pi^i$ to handle deformable objects $\langle$\textbf{4}$\rangle$, such as humans, and to model appearance changes of objects in the training set $\langle$\textbf{5}$\rangle$. For every static object, we assume $x_t^i$ to be fixed and known. For playable object $i$ instead, given the current camera parameters and its bounding box $b_t^i$, we approximate $x_t^i$ by projecting the middle point of the lower bounding box edge onto the ground plane. We then compute the style and pose descriptors using a convolutional encoder network $E$ for each object. The encoder takes as input the image cropped at the location defined by the bounding box for each object and outputs both $w_t^i$ and $\pi_t^i$.
In the rest of the paper, we omit object indexes.



 We introduce a novel synthesis module detailed in Sec~\ref{sec:encoding_synthesis}. The action module is described in Sec.~\ref{sec:action module}. The training procedures are given in Secs.~\ref{sec:training-synthesis} and \ref{sec:training}.

\subsection{Synthesis Module}
\label{sec:encoding_synthesis} 

\begin{figure}[ht]
    \centering
    \includegraphics[width=0.90\columnwidth]{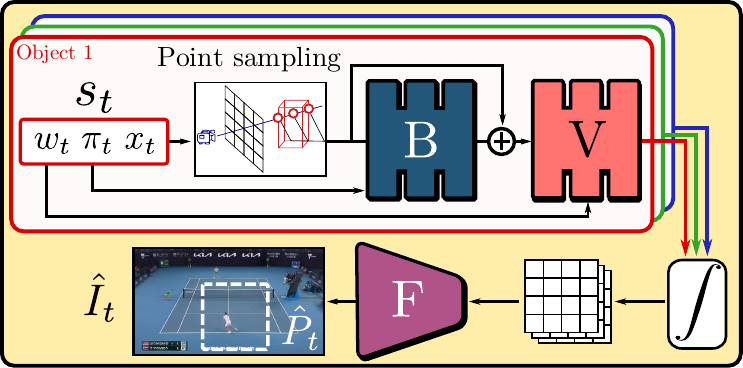}
    \caption{\textbf{The synthesis module} consists of two steps. First, non-rigid neural radiance fields with a bending network $B$ and style modulation are used to generate a feature map. Second, the feature maps are fed to a ConvNet $F$.} 
    \label{fig:nerf}
\end{figure}

%
The aim of the synthesis module is to reconstruct the input image from the camera pose and {states} $s_t$.
We found NeRF \cite{mildenhall2020nerf} to be a {reasonable} base architecture for explicit camera control $\langle$\textbf{2}$\rangle$. 
Therefore, we propose a novel architecture (Fig.~\ref{fig:nerf}) that combines non-rigid neural radiance fields with a convolutional image generator to address $\langle$\textbf{2}-\textbf{6}$\rangle$. 

\noindent\textbf{Camera Control $\langle$2$\rangle$} is achieved by employing NeRF \cite{mildenhall2020nerf} as a base architecture. Our NeRF represents scenes using a fully-connected network $V$, whose input is a single vector containing a point location in 3D. 
It outputs the volume density $\sigma$ and radiance $c$ for the input point location. Given a desired virtual camera, 3D points are sampled along the camera ray $r$ traced through each pixel. The color value of every pixel is computed by integration over the ray $r$:
\begin{equation}
    C(r) = \int_{t_n}^{t_f} e^{-\int_{t_n}^{t} \sigma(r(s)) ds}\sigma(r(t))c(r(t)) dt \label{eq:nerf}
\end{equation}


Similarly to \cite{niemeyer2021giraffe,hao2021GANcraft}, instead of directly predicting color values, our neural radiance fields generate feature maps for the input camera pose, while a convolutional image generator is in charge of generating realistic frames. 
For more details about NeRFs, please refer to the \emph{Supp. Mat.} and\cite{mildenhall2020nerf}.

\noindent\textbf{Multi-object $\langle$3$\rangle$.} 
Each object is modeled using a separate feature field parametrized as an object-specific MLP $V$. The field is bounded by volume $\beta$ and centered at the respective object location $x_t$. Given a ray $r$, we compute its features $f(r)$ according to the following procedure. We first intersect $r$ with each bounding volume $\beta$ to compute the ingress and egress location of the ray with each object $x_{\mathrm{in}}$, $x_{\mathrm{out}}$. For each object, we then uniformly sample a given amount of positions $\{x_p\}_{p=1}^{N}$ between $x_{\mathrm{in}}$ and $x_{\mathrm{out}}$ and obtain the respective features $f_p$ and opacities $\sigma_p$ as $f_p, \sigma_p = V(x_p)$. $f(r)$ is obtained by integration similarly to Eq.~\eqref{eq:nerf}. 

\noindent\textbf{Deformable objects  $\langle$4$\rangle$.} To handle deformable objects such as humans, we make use of non-rigid NeRF models, similarly to \cite{tretschk2021nonrigid}. 
For each playable object, we introduce a ray bending network $B$ parametrized as an MLP. Given an object pose descriptor $\pi$ and position $x_p$ on ray $r$, we use the bending network to regress the corresponding position $\tilde{x}_p$ on the bent ray $\tilde{r}$ as:
\begin{equation}
    \tilde{x}_p = x_p + B(x_p, \pi_t)
\end{equation}
We then make use of the positions on $\tilde{r}$ when sampling $V$. In this way, $B$ encodes the transformation from the space of the deformed object to a canonical space and $V$ encodes a canonical representation of the object.

\noindent\textbf{Appearance changes $\langle$5$\rangle$.} 
The appearance of each object may vary widely in the dataset. In order for each object-specific model to be able to represent the complete set of possible appearances of its object, we propose the use of a style embedding layer inspired by AdaIN \cite{huang2017adain}, which we embed into $V$. Assuming a hidden feature $h_t$ in $V$ and a style code $w_t$, we modulate $h_t$ as follows:
\begin{equation}
    \tilde{h}_t = \gamma(w_t) h_t + \beta(w_t)
\end{equation}
where $\gamma$ and $\beta$ are trainable linear layers. Following \cite{mildenhall2020nerf}, we design $V$ as a backbone terminated by two separate branches, one for opacity and one for features prediction. We assume that the style of an object should influence its features, but not its geometry. Therefore, we insert our modulation layer in the features prediction branch only. 

\noindent\textbf{Robustness $\langle$6$\rangle$} to calibration and localization errors is achieved through a Feature Renderer. Our compositional NeRF model outputs a feature map $f_t$ corresponding to an input image patch. We employ a ConvNet $F$ to reconstruct it. Due to the ability of ConvNets to model cross-pixel relationships, inaccuracies in the estimation of features caused by input noise can be compensated, reducing the associated blur. 
Note that $F$ contains upsampling layers. It allows an important reduction in the number of rays that are to be sampled by the NeRF model since it outputs a feature map at a lower resolution than the image. Therefore, we reduce memory consumption allowing larger patches to be rendered. We also find it beneficial to use multiple input feature maps at different resolutions to capture details at different scales (see \emph{Sup. Mat.} for details).


\subsection{Action Module}
\label{sec:action_module}
\begin{figure}[t]
    \centering
    \includegraphics[width=0.70\columnwidth]{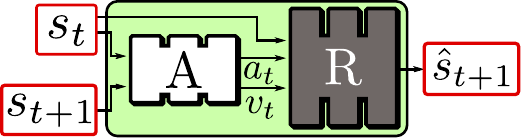}
    \caption{\textbf{The action module.} Given the states at times $t$ and $t\!+\!1$, the action network $A$ predicts a discrete action label $a_t$ and action variability $v_t$ that are combined by the dynamics network $R$ to estimate the new environment state $s_{t+1}$ given the old $s_t$.} 
    \label{fig:action}
\end{figure}

\label{sec:action module}

The action module (Fig.~\ref{fig:action}) learns the action space and enables playability \textbf{$\langle$1$\rangle$}. The actions of each playable object are modeled by a separate action module, consisting of the action and dynamics networks.



\noindent\textbf{Action Network.}
Given two successive {environment states} $s_t$ and $s_{t+1}$, we use an action network $A$ to infer a discrete representation $a_t \in \{1,..., K\}$ of the action performed by the object in the input sequence. Following \cite{menapace2021pvg}, to address non determinism present in the environment, we also extract an action variability embedding $v_t$ describing the particular variation of $a_t$ performed at time $t$:
\begin{equation}
    a_t, v_t = A(s_t, s_{t+1})
\end{equation}

\noindent\textbf{Dynamics Network.}
The role of the dynamics network is to predict the state $s_{t+1}$ from $s_t$ and the action label $a_t$. We adopt a recurrent model $R$ implemented as an LSTM to model the dynamics of the object. The next state prediction $\hat{s}_{t+1}=(\hat{x}_t, \hat{w}_t, \hat{\pi}_t)$ is given by:
\begin{equation}
    \hat{s}_{t+1} = R(s_t, a_t, v_t)\label{eq:dynamics}
\end{equation}
In our preliminary experiments, we observe that when R directly regresses $\hat{x}_{t+1}$ as formulated in \eqref{eq:dynamics}, the model learns actions that are independent from the current camera position. This behavior is unnatural for the user since in applications such as video-games, object movements are typically expressed relatively to the camera pose. To avoid this behavior, $R$ is instead asked to predict the object movement $\Delta$ expressed in the camera coordinate system. The estimated position is then given by $\hat{x}_{t+1} = x_{t} + M\Delta$ where $M$ is the rotation matrix expressing the orientation of the camera.

\vspace{-1mm}
\subsection{Synthesis Module Training}
\label{sec:training-synthesis}
We train our model in two steps by first training the encoder and synthesis module until convergence, and then the action module. We train the encoder and synthesis module using the perceptual loss of Johnson \etal \cite{johnson2016perceptual} that assesses image reconstruction quality in features spaces of a pretrained VGG network. The loss is computed between the ground truth and reconstructed image patches.  The perceptual loss is complemented by an L2 reconstruction loss in the pixel space. 

Our preliminary experiments showed that training may fail to correctly disentangle object style and pose (\ie$w$ and $\pi$ respectively). 
Indeed, the reconstruction losses can be minimized using $w$ alone by predicting a constant, non-deforming surface with changing style. To avoid this problem, we make the observation that the pose of an object in neighboring frames can change while the style doesn't. Therefore, we enforce better disentanglement by permuting the order of $w$ codes along the temporal dimension for each sequence before feeding them to the synthesis module.


\vspace{-1mm}
\subsection{Action Module Training} 
\label{sec:training}
In the second phase of training, we train the action module using a combination of losses. Each loss is computed separately for each playable object and then averaged to produce the final optimization objective.

\noindent\textbf{Reconstruction loss}.
For each playable object, we reconstruct the input sequence of {environment states} $\{s_t\}_{t=1}^T$, obtained by encoding each input image using the encoder $E$, and impose an $\ell_2$ reconstruction loss $\mathcal{L}_\mathrm{rec}$ with the corresponding reconstructed sequence $\{\hat{s}_t\}_{t=1}^T$.

\noindent\textbf{Action learning losses}.
We employ the information-theoretic action learning loss of \cite{menapace2021pvg} to foster the understanding of actions. For each playable object, the action network $A$ produces internal estimates of action probabilities $p_t$ and $\hat{p}_t$ for input $s_t$ and reconstructed $\hat{s}_t$ {environment states} respectively. By imposing the maximization of mutual information between these two distributions we foster the action network both to discover the $K$ action categories, avoiding mode collapse, and to produce consistent action estimates for the input and reconstructed sequence:
\begin{align}
  \mathcal{L}_\mathrm{act} = -\mathcal{MI}(p_t,\hat{p}_t).
\end{align}
\vspace{-1mm}
In addition, to improve consistency between discrete actions and 3D movements, we propose to optimize a novel loss consisting in a soft version of the $\Delta$ Mean Squared Error ($\Delta$-MSE) introduced in \cite{menapace2021pvg}. This metric is based on the idea that same actions $a_t$ should correspond to similar object motions $\Delta$.   
Assuming a batch containing $J$ image pairs, we extract the object motion $\Delta_j, j\!\in\!\{1,...,J\}$ and estimate the mean object motion for each action:
\begin{equation}
  \forall k\in \{1,...,K\},   \mu_k = \frac{\sum_{j=1}^{J} p_{jk} \Delta_{j}}{\sum_{j=1}^{J} p_{jk}}
\end{equation}
where $p_{jk}$ denotes the probability for the image pair $j$ to be assigned to the action $k$. We then minimize the mean squared distance between the motion $\Delta_j$ and the mean motion for each action:
\begin{equation}
    \mathcal{L}_{\Delta} = \frac{1}{\mathrm{Var}(\Delta)}\sum_{j=1}^{J}\sum_{k=1}^{K} p_{jk} \left\|\Delta_j - \mu_k\right\|_2^2
\end{equation}
where $\mathrm{Var}(\Delta)$ is used as a normalization factor.

\noindent\textbf{Temporal Discriminator.}
Previous methods for playable video generation \cite{menapace2021pvg}, tend to produce sequences where the playable objects move in the scene with unrealistic motions. We attribute this behavior to the use of reconstruction as the main training objective. 
Optimizing reconstruction losses does not penalize action representations that lead to temporally inconsistent videos.  
To address the problem, for each playable object we introduce a temporal discriminator $D$ implemented as a 1D ConvNet over the temporal dimension. Given a sequence of {environment states}, the temporal discriminator is trained to classify them as real if produced by encoding the input images using $E$ or as fake if reconstructed by the action module. We implement our adversarial training procedure using a vanilla GAN loss with loss terms $\mathcal{L}_{\mathrm{G}}$ and $\mathcal{L}_{\mathrm{D}}$ for the action module and temporal discriminator respectively.


\noindent\textbf{Total loss}.
Our optimization objective for $A$ and $R$ is
\begin{equation}
    \mathcal{L} = \lambda_\mathrm{rec}\mathcal\mathcal{L}_\mathrm{rec} + \lambda_\mathrm{act}\mathcal{L}_\mathrm{act} + \lambda_{\Delta}\mathcal{L}_{\Delta} + \lambda_{\mathrm{G}}\mathcal{L}_{\mathrm{G}}
\end{equation}
where we introduce the weighting parameters $\lambda_\mathrm{rec}$, $\lambda_\mathrm{act}$, $\lambda_{\Delta}$ and $\lambda_{\mathrm{G}}$. For training $D$, we minimize the adversarial objective $\mathcal{L}_{\mathrm{D}}$ of the discriminator.

\noindent\textbf{Inference.} 
At inference time, we assume that only the first frame of the sequence is given. We use the encoder module to extract the first {environment state} $\hat{s}_1 = s_1$. At each timestep $t$, we let the user specify a discrete action for each playable object and use the dynamics network $R$ to derive $\hat{s}_{t+1}$ in an autoregressive way. Since the action input is specified by the user, during inference we do not make use of the action network and always set $v_t=0$. The environment states generated by the dynamics network are rendered to images using the synthesis module. 


                  
\section{Experiments}

\noindent\textbf{Datasets.} Evaluating $\langle$\textbf{1}-\textbf{6}$\rangle$ is challenging and requires video datasets featuring camera motion $\langle$\textbf{2}$\rangle$, multiple playable objects $\langle$\textbf{1},\textbf{3}$\rangle$, deforming objects $\langle$\textbf{4}$\rangle$ and varied appearance $\langle$\textbf{5}$\rangle$. For this reason, we collect three datasets:\\
\noindent \textbullet~\textit{Minecraft} dataset. We collect a synthetic video dataset with duration of 1h with two sparring \emph{Minecraft} \cite{minecraft} players. Wide camera movement and diverse, deforming players allow the evaluation of $\langle$\textbf{1}$-$\textbf{5}$\rangle$.\\ 
\noindent \textbullet~\textit{Minecraft Camera} dataset. We collect \emph{Minecraft} \cite{minecraft} sequences where the camera is moved in the neighborhood of a starting position. We use these frames as a synthetic ground truth for the evaluation of camera control $\langle$\textbf{2}$\rangle$.\\
\noindent \textbullet~\textit{Tennis} dataset. We collect a large-scale dataset of 43 broadcast tennis matches totalling 12h of videos for the evaluation of $\langle$\textbf{1}-\textbf{6}$\rangle$. The dataset features challenging player poses $\langle$\textbf{5}$\rangle$, high variability in tennis fields and players $\langle$\textbf{4}$\rangle$ and noise in camera estimation and player localizaton $\langle$\textbf{6}$\rangle$.\\
To allow comparison with playable video generation methods under their simplifying assumptions, we adopt the \emph{Tennis} dataset of \cite{menapace2021pvg}, referred to as \textit{Static Tennis}. The dataset features limited camera movement, each video is cropped to depict only a single player, only one field is present and players have uniform appearance, thus only $\langle$\textbf{1},\textbf{4}$\rangle$ are evaluated.
The datasets are detailed in the \emph{Supp. Mat.}.

\noindent\textbf{Evaluation Protocol.} 
We perform a separate evaluation of the synthesis $\langle$\textbf{2}-\textbf{6}$\rangle$ and the action modules $\langle$\textbf{1}$\rangle$ using similar evaluation protocols. For the former, we reconstruct each test sequence by extracting the environment state of each frame and rendering the original frame back. For the action module, we follow the evaluation protocol of \cite{menapace2021pvg}. In particular, we consider a test sequence and extract the environment state of the first frame, then we use the action network to extract the sequence of discrete actions present in the sequence and reconstruct each frame starting from the first environment state. 
%
As video quality metrics $\langle$\textbf{2},\textbf{4}-\textbf{6}$\rangle$ we adopt \textit{LPIPS}~\cite{zhang2018unreasonable}, \textit{FID}~\cite{heusel2017advances} and \textit{FVD}~\cite{unterthiner2018towards} computed between the test sequences and the reconstructed sequences.
For evaluation of the action space $\langle$\textbf{1},\textbf{3}$\rangle$, following \cite{menapace2021pvg}, we define $\Delta$ as the difference in position of an object between two given frames and use the following metrics:\\
\noindent \textbullet~\textit{$\Delta$ Mean Squared Error ($\Delta$-MSE)}: The expected error in terms of MSE in the regression of $\Delta$ from a discrete action. For each action, the average $\Delta$ is used as the optimal estimator. The metric is normalized by the variance of $\Delta$.\\
\noindent \textbullet~\textit{$\Delta$-based Action Accuracy ($\Delta$-Acc)}: The accuracy with which a discrete action can be regressed from $\Delta$.\\
\noindent \textbullet~\textit{Average Detection Distance (ADD)}: The average Euclidean distance between the bounding box centers of corresponding objects in the test and reconstructed frames.\\
\noindent \textbullet~\textit{Missing Detection Rate (MDR)}: The portion of detections that are present in the test sequences but that are not matched by any detection in the reconstructed sequences.

\subsection{Comparison on Playable Video Generation}

\begin{table}[t]
\begin{center}

\setlength{\tabcolsep}{1.0pt}
\footnotesize
\begin{tabular}{lccccccc}
\toprule
 & LPIPS$\downarrow$  & FID$\downarrow$ & FVD$\downarrow$& $\Delta$-\emph{MSE}$\downarrow$ & $\Delta$-\emph{Acc}$\uparrow$ & ADD$\downarrow$ & MDR$\downarrow$\\
\midrule
MoCoGAN \cite{tulyakov2018moco} & 0.266 & 132 & 3400 & 101 & 26.4 & 28.5 & 20.2 \\
MoCoGAN+ & 0.166 & 56.8 & 1410 & 103 & 28.3 & 48.2 & 27.0\\
SAVP \cite{lee2018savp} & 0.245  & 156 & 3270 & 112 & 19.6 & 10.7 & 19.7\\
SAVP+ & 0.104 & 25.2 & \textbf{223}& 116 & 33.1 & 13.4 & 19.2  \\
CADDY \cite{menapace2021pvg} & 0.102  & \textbf{13.7} & 239 & 72.2 & 45.5 & \textbf{8.85} & 1.01 \\
\midrule

(Ours) & \textbf{0.089}  & 15.3 & 237 & \textbf{32.8} & \textbf{68.1} & 9.47 & \textbf{0.15}\\

\bottomrule
\end{tabular}
\end{center}
\vspace{-8pt}
\caption{Comparison with PVG state of the art on the \textit{Static Tennis} dataset of \cite{menapace2021pvg}. $\Delta$-\emph{MSE}, $\Delta$-\emph{Acc} and MDR in \%, ADD in pixels.}
\vspace{-2mm}

\label{table:playability_old_tennis}
\end{table}

In this section, we evaluate the action-modeling capabilities of our method by comparing against the state of the art in the related problem of Playable Video Generation (PVG) \cite{menapace2021pvg} where the objective is to learn a set of discrete action labels in an unsupervised fashion to condition video generation. Differently from our setting, in PVG no explicit camera control is required. Moreover, existing PVG methods assume a single user-controllable object and that camera motion is limited. 

To satisfy these simplifying assumptions, we adopt the \emph{Static Tennis} dataset of \cite{menapace2021pvg}. Tab.~\ref{table:playability_old_tennis} shows the results. Our method substantially improves the $\Delta$-MSE and $\Delta$-Acc action quality metrics suggesting that the learned actions are better correlated with player movement. In addition, the reduced LPIPS and MDR indicate an improvement in the quality of the generated reconstruction which is supported by a user study shown in the \emph{Supp. Mat.}. We report qualitative results in the \emph{Supp. Mat.}.

\begin{table*}[ht]
\begin{center}

\setlength{\tabcolsep}{1.0pt}
\footnotesize
\begin{tabular}{lccc|ccccccc|cccc}
\toprule
\multicolumn{4}{c}{}  & \multicolumn{7}{c}{\emph{Tennis}} & \multicolumn{4}{c}{\emph{Minecraft Camera}} \\
\midrule
 & \emph{Aux.} & \emph{H.Res.} & $\mathcal{L}_{\Delta}$ & LPIPS$\downarrow$ & FID$\downarrow$ & FVD$\downarrow$ & $\Delta$-\emph{MSE}$\downarrow$ & $\Delta$-\emph{Acc}$\uparrow$ & ADD$\downarrow$ & MDR$\downarrow$ & LPIPS$\downarrow$ & FID$\downarrow$ & ADD$\downarrow$ & MDR$\downarrow$\\
\midrule
CADDY \cite{menapace2021pvg} (i) &&&& 0.313 & 61.0 & 877 & 0.901 & 42.6 & 35.1 & 36.9 & 0.747 & 306 & 11.7 & 95.8 \\
CADDY \cite{menapace2021pvg} (ii) &\checkmark&&& 0.351 & 69.2 & 1109 & 0.592 & 59.6  & 29.0 & 24.8 & 0.762 & 324 & 44.7 & 92.2 \\
CADDY \cite{menapace2021pvg} (iii) &\checkmark&\checkmark&& 0.213 & \textbf{15.4} & 727 & 0.693 & 57.5  & 18.7 & 11.7 & 0.669 & 244 & 29.2 & 82.0 \\
CADDY \cite{menapace2021pvg} (iv) &\checkmark&&\checkmark& 0.445 & 70.3 & 1568 & 0.797 & 62.4 & 29.6 & 33.0 & 0.699 & 314 & 62.0 & 89.4 \\
CADDY \cite{menapace2021pvg} (v) &\checkmark&\checkmark&\checkmark& 0.534 & 191 & 8083 & 0.633 & 73.5  & 20.2 & 60.3 & 0.679 & 337 & 19.1 & 93.6 \\

\midrule

(Ours) &&&& \textbf{0.181} & 17.4 & \textbf{485} & \textbf{0.293} & \textbf{95.7} & \textbf{14.0} & \textbf{4.84} & \textbf{0.242} & \textbf{29.2} & \textbf{5.69} & \textbf{8.07} \\

\bottomrule
\end{tabular}
\end{center}
\vspace{-8pt}
\caption{Playability evaluation with baselines on the \emph{Tennis} dataset and camera control evaluation on the \emph{Minecraft Camera} dataset. \emph{Aux.}: use of auxiliary bounding box and camera pose information; \emph{H.Res.} use of the high resolution model; $\mathcal{L}_{\Delta}$ use of the loss for $\Delta$-\emph{MSE}. $\Delta$-\emph{MSE}, $\Delta$-\emph{Acc} and MDR in \%, ADD in pixels.}
\vspace{-5mm}
\label{table:playability_tennis_minecraft_camera}
\end{table*}

\subsection{Comparison with Previous Methods}

\noindent\textbf{Baselines.} We propose to build baselines for the creation of PEs from state of the art methods in the related problem of PVG. We make use of the following set of versions of CADDY \cite{menapace2021pvg} which are modified to account for multiple playable objects and for camera motion: (i) the action network produces a distinct output for each dynamic object in the environment; (ii) (i) + the action and dynamics networks are conditioned on bounding box and camera information; (iii) (ii) + output resolution is increased to match our method; (iv) (ii) + $\mathcal{L}_{\Delta}$; (v) (iii) + $\mathcal{L}_{\Delta}$.

\begin{figure}[]
    \centering
    \includegraphics[trim={0 0mm 0 0mm},clip,width=\columnwidth]{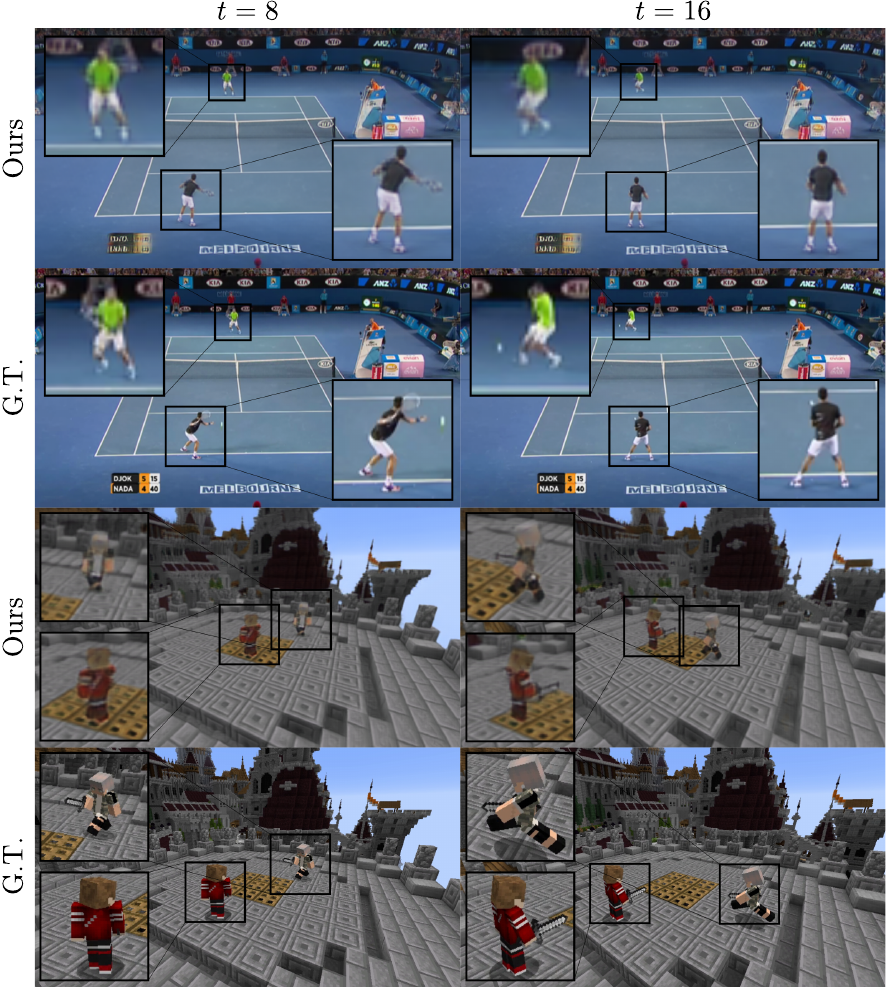}
    \captionof{figure}{Qualitative reconstruction results produced by our method on the \emph{Tennis} and \emph{Minecraft} datasets. In the reconstructed sequence, playable objects move according to the ground truth sequence and are rendered in realistic poses.}
    \label{fig:playability_reconstruction_tennis_minecraft}
    \vspace{-3mm}
\end{figure}

\begin{figure}
    \centering
    \includegraphics[width=\columnwidth]{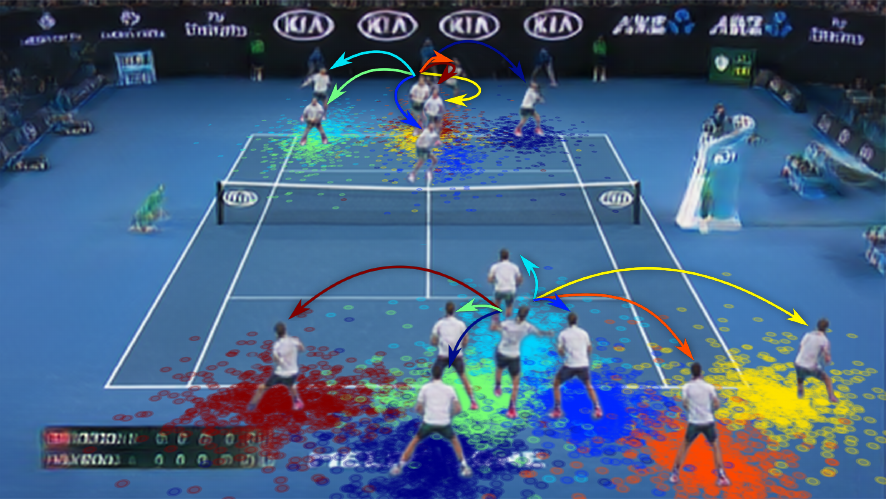}

    \caption{Action space learned by our method on the \emph{Tennis} dataset. Each color represents a learned action and each arrow shows the effects of applying the respective action six times to the initial player. The overlay on the floor shows the distribution of possible ending positions after the application of each action.}
    \label{fig:action_qualitatives}
\end{figure}

\noindent\textbf{Playability evaluation $\langle$\textbf{1}$\rangle$.}
We evaluate player control capabilities in Tab.~\ref{table:playability_tennis_minecraft_camera} and in the \emph{Supp.~Mat.}.
On the \emph{Tennis} dataset our model substantially improves over the baselines in the action space metrics, LPIPS and FVD, suggesting better controllability of the players. In particular, the considerably lower MDR indicates a better capacity of the model in generating players with respect to the baselines. Fig.~\ref{fig:playability_reconstruction_tennis_minecraft} shows qualitative reconstruction results for our method. As suggested by the MDR and ADD scores, the model correctly synthesizes both players and is able to reconstruct the player movements of the ground truth sequence using only a sequence of discrete actions. In addition, a visualization of the learned action space (see Fig.~\ref{fig:action_qualitatives}) shows that the model learns a set of diverse discrete actions that correspond to the main movement directions.

To further evaluate the quality of the action space, we perform a user study (see \emph{Supp. Mat.}) on the \emph{Tennis} dataset, following the protocol of Menapace \etal~\cite{menapace2021pvg}. To evaluate the consistency of learned actions, we measure user agreement using the Fleiss' kappa measure \cite{fleiss1971measuring}. Our method achieves an agreement of 0.444, while the best baseline shows a lower agreement of 0.353.


\noindent\textbf{Camera control evaluation $\langle$\textbf{2}$\rangle$.}
We evaluate the quality with which the model can synthesize novel views. We choose to perform a quantitative evaluation on the \emph{Minecraft Camera} dataset since novel view ground truth is present. We start from the first frame and reconstruct each sequence using the camera parameters of the novel views. Results are shown in Tab.~\ref{table:playability_tennis_minecraft_camera}. Despite the presence of auxiliary bounding box and camera pose inputs for CADDY \cite{menapace2021pvg}, the baseline method fails in synthesizing the scene from novel perspectives. We ascribe this phenomenon to the lack of an explicit model for the camera. Our method instead can successfully synthesize the scene from novel camera perspectives.

\begin{figure}
     \centering

     \includegraphics[width=1.0\linewidth]{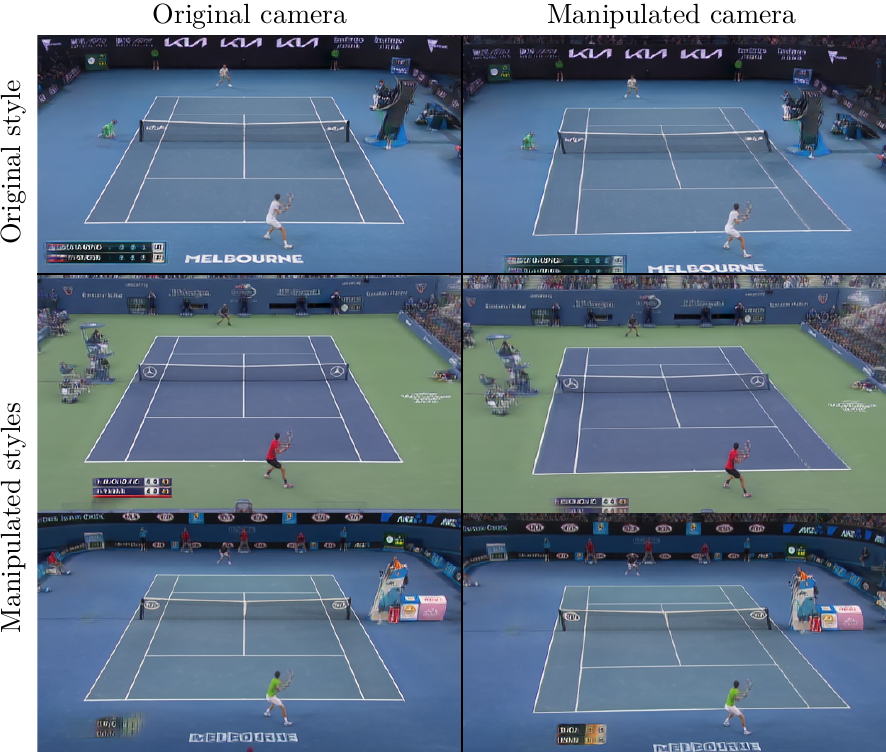}
     \caption{Camera and style manipulation results on the \emph{Tennis} dataset. The original image is rendered under a novel camera perspective using varying styles for the field and players.}
    \vspace{-1mm}
    \label{fig:qualitatives_style_camera}
\end{figure}

In Fig.~\ref{fig:qualitatives_style_camera} we show qualitative camera and style manipulation results for our method on the \emph{Tennis} dataset. Our model can synthesize the scene under novel views and correctly alter the style of the field and players to the one of a target image. We present additional camera and style manipulation results in the \emph{Supp. Mat.}.

\begin{table}
\begin{center}
\setlength{\tabcolsep}{1.8pt}
\footnotesize
\begin{tabular}{lccccccccc}
\midrule
Var. & \emph{Multi}\,$\langle$\textbf{3}$\rangle$ & $\pi$\,$\langle$\textbf{4}$\rangle$ & $w$\,$\langle$\textbf{5}$\rangle$ & $F$\,$\langle$\textbf{6}$\rangle$ & LPIPS$\downarrow$ & FID$\downarrow$ & FVD$\downarrow$  & ADD$\downarrow$ & MDR$\downarrow$ \\
\midrule
(a) &&&&& 0.735 & 376 & 2548 & 109.1 & 99.9 \\
(b) &\checkmark&&&& 0.595 & 266 & 1617 & 45.4 & 86.4 \\
(c) &\checkmark&\checkmark&&& 0.648 & 301 & 1818 & 10.17 & 50.2 \\
(d) &\checkmark&\checkmark&$\sim$&& 0.361 & 68.6 & 482 & 7.39 & 31.9 \\
(e) &\checkmark&\checkmark&\checkmark&& 0.350 & 61.0 & 465 & 8.27 & \textbf{31.8} \\
(f) &\checkmark&\checkmark&\checkmark&$\sim$& 0.341 & 67.4 & 1371 & 88.5 & 88.8 \\
\midrule

Full &\checkmark&\checkmark&\checkmark&\checkmark& \textbf{0.193} & \textbf{16.5} & \textbf{289} & \textbf{5.45} & 33.7 \\
\midrule
\end{tabular}
\end{center}
\vspace{-8pt}
\caption{Synthesis module ablation results on the \textit{Minecraft} dataset. \emph{Multi}: use of multi-object modeling, $\pi$: use of deformation, $w$: use of style modulation layers or of direct style encoding ($\sim$), $F$: use of the feature renderer or of the simplified renderer ($\sim$). ADD in pixels, MDR in \%.}
\vspace{-2mm}

\label{table:ablation_reconstruction_minecraft}
\end{table}

\subsection{Ablation Studies}


\noindent\textbf{Synthesis module Ablation Study $\langle$\textbf{3}-\textbf{6}$\rangle$.}
In this section we evaluate the contribution of each proposed architecture component for the synthesis module: \emph{Multi} use of multi-object modeling $\langle$\textbf{3}$\rangle$, $\pi$ use of deformation modeling $\langle$\textbf{4}$\rangle$, $w$ use of style modulation layers for appearance changes $\langle$\textbf{5}$\rangle$, $F$ use of the feature renderer for robustness $\langle$\textbf{6}$\rangle$.
We produce the following method variations: (a) no component is used; this approach resembles NeRF \cite{mildenhall2020nerf}; (b) \emph{Multi}; (c) \emph{Multi} and $\pi$; this architecture is akin to NR-NeRF \cite{tretschk2021nonrigid} with $\langle$\textbf{3}$\rangle$; (d) \emph{Multi}, $\pi$, and $w$ injected with concatenation rather than style modulation layers; (e) \emph{Multi}, $\pi$, and $w$ with style modulation layers; (f) \emph{Multi}, $\pi$, $w$ and a simplified ConvNet $F$ that renders the complete frame from feature maps at a single resolution; from an architectural viewpoint, this feature rendering strategy resembles the one of GIRAFFE \cite{niemeyer2021giraffe}.


Results are shown in Tab.~\ref{table:ablation_reconstruction_minecraft} and in the \emph{Supp. Mat.}. (c) and (e) show that deformation and style modeling with style modulation layers are both necessary to accurately synthesize the scene, but generate blurry results due to calibration and localization errors. We recover sharpness by introducing our ConvNet feature renderer which reduces blur by modeling cross-pixel correlations. Substituting our renderer with the one of (f) leads to performance degradation due to the excessively sparse sampling of rays imposed by memory constraints when rendering the complete frame that leads to 3D consistency artifacts which are particularly apparent in the region of dynamic objects.


\noindent\textbf{Action module Ablation Study.}
We now evaluate the contribution of the main components of the action module by ablating the following: \emph{Rel.} use of camera-relative object movement in the dynamics network; $D$ use of the temporal discriminator; $\mathcal{L}_{\Delta}$ use of the loss on $\Delta$-MSE; $\mathcal{L}_\mathrm{act}$ use of the information-theoretic action learning loss. Results are shown in Tab.~\ref{table:ablation_playability_minecraft_reduced}. Removing the temporal discriminator causes an increase in the FVD. A qualitative analysis of the results (see \emph{Supp. Mat.}) shows that models not using $D$ produce sequences where the players translate in the scene, but fail to realistically move their limbs. In addition, the introduction of $\mathcal{L}_{\Delta}$ produces a positive impact on the action space metrics. We also note that, thanks to the presence of $\mathcal{L}_{\Delta}$, the model learns an action space even in the absence of  $\mathcal{L}_\mathrm{act}$. Lastly, without camera-relative object movement in the dynamics network, the model produces movements that are independent from the current camera orientation, which is undesirable (Sec.~\ref{sec:action module}). 

\begin{table}
\begin{center}

\setlength{\tabcolsep}{1.0pt}
\footnotesize
\begin{tabular}{lccccccccccc}
\toprule
Var. & \emph{Rel.} & $D$ & $\mathcal{L}_{\Delta}$ & $\mathcal{L}_\mathrm{act}$ &  LPIPS$\downarrow$ & FID$\downarrow$ & FVD$\downarrow$ & $\Delta$-\emph{MSE}$\downarrow$ & $\Delta$-\emph{Acc}$\uparrow$ & ADD$\downarrow$ & MDR$\downarrow$\\
\midrule
(A) &&&&\checkmark& 0.205 & 17.0 & 334 & 0.903 & 33.9 & 18.7 & \textbf{33.0} \\
(B) &\checkmark&&&\checkmark& \underline{0.204} & 17.0 & \underline{329} & 0.290 & 76.0 & 18.6 & \underline{33.5} \\
(C) &\checkmark&&\checkmark&\checkmark& \textbf{0.203} & \underline{16.9} & 340 & \textbf{0.263} & \textbf{80.0} & \textbf{15.4} & 34.0 \\
(D) &\checkmark&\checkmark&&\checkmark& \underline{0.204} & 17.0 & \textbf{323} & 0.289 & 77.0 & 17.8 & 34.3 \\
(E) &\checkmark&\checkmark&\checkmark&& \underline{0.204} & \underline{16.9} & 335 & 0.276 & 77.5 & \underline{17.5} & 34.0 \\

\midrule

Full &\checkmark&\checkmark&\checkmark&\checkmark& \underline{0.204} & \textbf{16.8} & \underline{329} & \underline{0.271} & \underline{77.7} & 17.8 & 33.9 \\

\bottomrule
\end{tabular}
\end{center}
\vspace{-8pt}
\caption{Action module ablation results on the \textit{Minecraft} dataset. \emph{Rel.} use of camera relative residual $\Delta$ output, $D$ use of the temporal discriminator, $\mathcal{L}_{\Delta}$ use of the loss for $\Delta$-\emph{MSE}, $\mathcal{L}_\mathrm{act}$ use of the information-theoretic action learning loss. $\Delta$-\emph{MSE}, $\Delta$-\emph{Acc} and MDR in \%, ADD in pixels.}
\vspace{-1mm}
\label{table:ablation_playability_minecraft_reduced}
\end{table}

\section{Conclusions and Discussion}

In conclusion, we present a new framework featuring a NeRF-based encoder-decoder architecture and an action module for the creation of compelling playable environments. Extensive experimental evaluation on two large-scale datasets shows that our method achieves state of the art performance. We discuss the main limitations and ethical aspects of the method in the \emph{Supp. Mat.}.


\section{Acknowledgements}

VG and CT  were supported by the ERC consolidator grant 4DReply (770784). This project was supported by the EU H2020 project AI4Media (951911).

{\small
\bibliographystyle{ieee_fullname}
\bibliography{egbib}
}

\end{document}